\documentclass[letterpaper, 10 pt, conference]{ieeeconf}  

\IEEEoverridecommandlockouts                              

\usepackage{tabularx}
\usepackage{amsmath,amssymb,amsfonts}
\usepackage[dvipsnames]{xcolor}
\usepackage{arydshln}
\usepackage{graphicx}
\usepackage{float}
\usepackage{caption}
\usepackage{circledsteps}
\usepackage{helvet}
\usepackage[normalem]{ulem}
\usepackage{hyperref}

\title{\LARGE \bf
Autonomous Robotic Pepper Harvesting:\\Imitation Learning in Unstructured Agricultural Environments}

\author{Chung Hee Kim, Abhisesh Silwal, George Kantor
\thanks{C. H. Kim, A. Silwal and G. Kantor are with the Robotic Institute at Carnegie Mellon University, Pittsburgh PA, USA \texttt{\{chunghek, asilwal, gkantor\}@andrew.cmu.edu}}
\thanks{Our project is open-sourced at \url{https://kantor-lab.github.io/roboharvest/}}
}

\begin{document}

\maketitle
\thispagestyle{empty}
\pagestyle{empty}

\begin{abstract}Automating tasks in outdoor agricultural fields poses significant challenges due to environmental variability, unstructured terrain, and diverse crop characteristics. We present a robotic system for autonomous pepper harvesting designed to operate in these unprotected, complex settings. Utilizing a custom handheld shear-gripper, we collected 300 demonstrations to train a visuomotor policy, enabling the system to adapt to varying field conditions and crop diversity. We achieved a success rate of 28.95\% with a cycle time of 31.71 seconds, comparable to existing systems tested under more controlled conditions like greenhouses. Our system demonstrates the feasibility and effectiveness of leveraging imitation learning for automated harvesting in unstructured agricultural environments. This work aims to advance scalable, automated robotic solutions for agriculture in natural settings.
\end{abstract}



\section{Introduction}


In light of rising food security concerns and a rapidly growing global population, the agricultural industry faces unprecedented pressure to enhance productivity and sustainability. Automation, particularly through robotics, has emerged as a critical pathway to meet these demands, helping to address labor shortages, imprecise environment monitoring, and inconsistent crop handling \cite{machines11010048}. 
Traditional farming, characterized by labor-intensive processes and variable efficiency, is being augmented by advanced technologies including robotics and artificial intelligence, offering potential solutions that can increase yield and promote sustainability in food production.

Among these advancements, the focus in robotic manipulation has shifted from traditional control algorithms to adaptive, learning-based approaches such as imitation learning. Imitation learning enables control policies to capture delicate behaviors observed in human demonstrations, which is particularly advantageous in agricultural tasks that demand complex physical interactions, such as harvesting and pruning. Traditional control algorithms often struggle to adapt to the natural variability in agricultural products and field conditions, while imitation learning enables robots to facilitate the precise manipulation needed in real world agricultural scenarios, substantially diminishing reliance on explicit programming \cite{mahmoudi11leveraging}. 

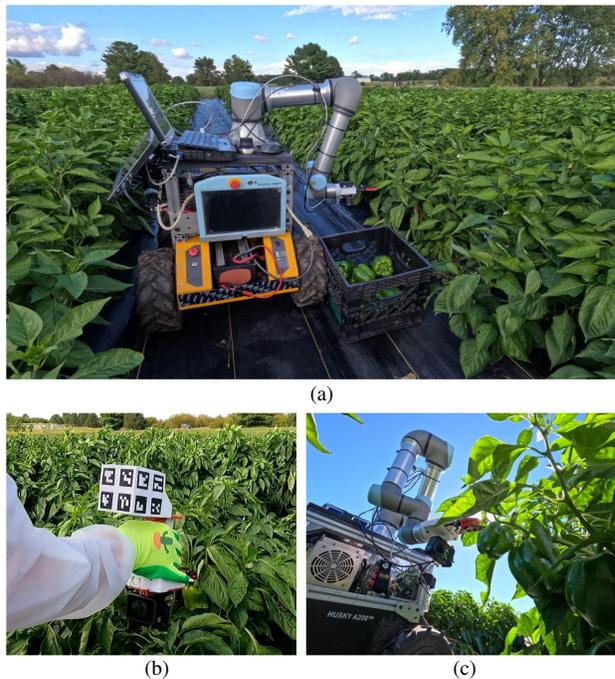
\begin{figure}[htbp]
\centering 
\includegraphics[width=0.95\columnwidth]{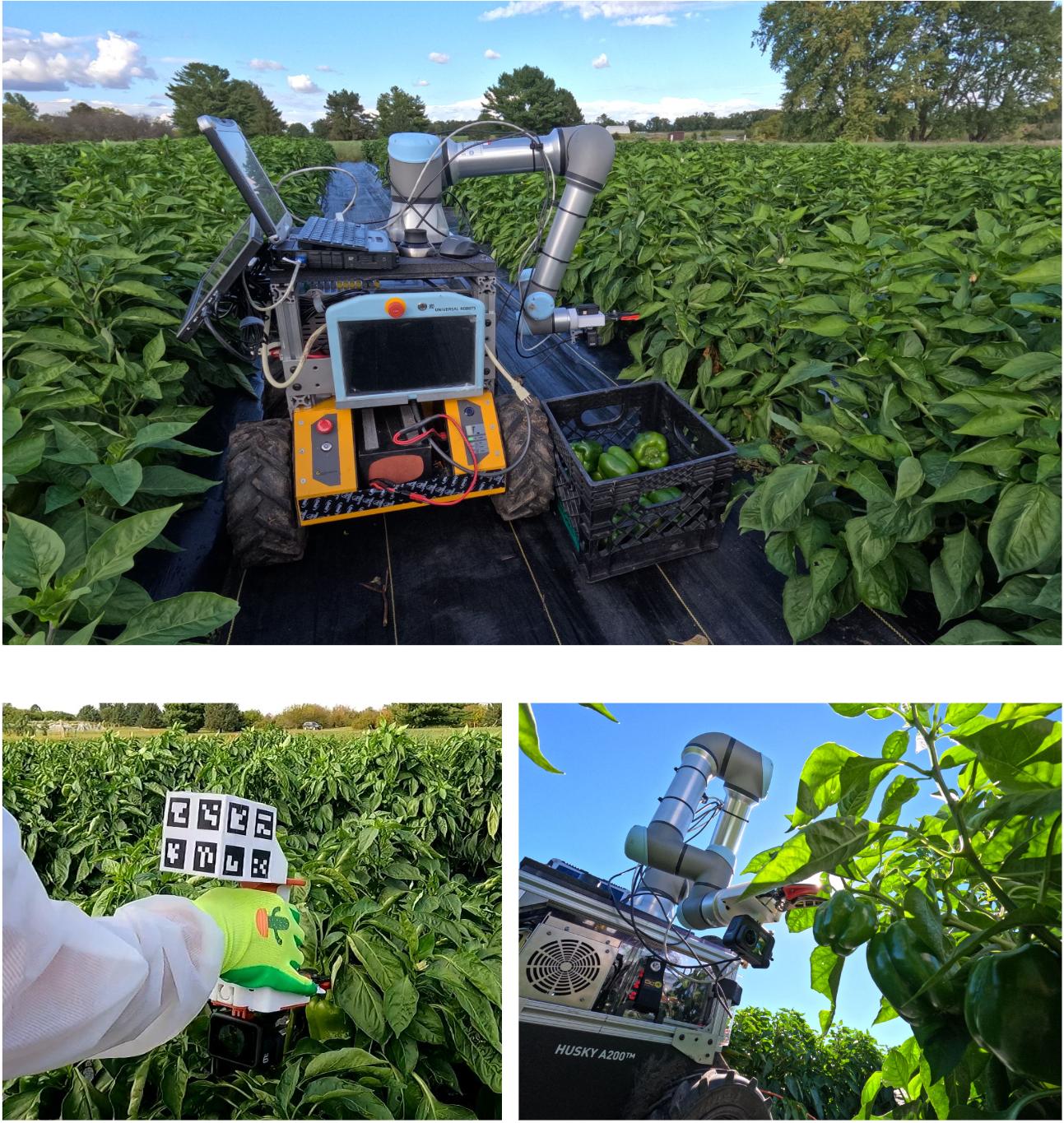}
\setlength{\unitlength}{1cm}
\begin{picture}(0,0)
\put(-4.4, 3.4){\footnotesize (a)}
\put(-6.6, -0.25){\footnotesize (b)}
\put(-2.5, -0.25){\footnotesize (c)}
\end{picture}
\vspace{4pt}
\caption{(a) Robotic automation in agriculture faces unique domain-specific challenges, including variable plant morphology, unpredictable lighting, and unstructured field conditions. (b) Pepper harvesting demonstrations are collected with a custom handheld shear-gripper device to train a visuomotor policy via imitation learning. (c) The trained policy enables autonomous robotic pepper harvesting in an outdoor field setting.}
\label{fig:main} 
\vspace{-20pt}
\end{figure}

In this work, we address the challenges of deploying imitation learning-based robotic systems in outdoor agricultural environments (Fig.~\ref{fig:main}(a)), which are characterized by diverse plant morphologies, inconsistent lighting, and unstructured dynamic field conditions. We build upon an adaptable data collection framework \cite{chi2024universal}, employing a customized handheld device outfitted with sensors to capture high-quality data of pepper harvesting demonstrations (Fig.~\ref{fig:main}(b)). By leveraging this approach, we demonstrate the effectiveness of a diffusion policy \cite{chi2023diffusionpolicy} trained on 300 real-world harvesting demonstrations. 
Our contributions are as follows:
\begin{itemize} 
    \item The first comprehensive study on robotic pepper harvesting in unprotected outdoor field settings, with a thorough evaluation and discussion of system performance. 
    \item The application of an imitation learning approach on agricultural manipulation that generalizes across environmental and crop variability in complex, unstructured field conditions, addressing unique domain-specific challenges.
    \item A publicly available dataset comprising 300 demonstrations of pepper harvesting in natural conditions, contributing valuable real-world data to the field. 
\end{itemize}
Through these contributions, our work aims to advance the adaptability and effectiveness of robotic systems in agriculture, highlighting the potential for imitation learning to support sustainable solutions in challenging field environments.


\section{Related Work}

Advancements in robotics and deep learning is driving the development of automated systems in various aspects of agriculture, including crop inspection and monitoring, as well as tasks requiring physical interaction like harvesting and pruning \cite{fountas2020agricultural}. For example, autonomous systems have been developed to perform crop yield estimation \cite{freeman20233d}, health monitoring \cite{10494888}, and modeling tree crops \cite{10160650}, \cite{10611327}. Several robotic systems have been developed for pruning vines  \cite{silwal2022bumblebee} and fruit trees \cite{you2022precision}, as well as harvesting strawberries \cite{xiong2020autonomous}, tomatoes \cite{7759122}, and aubergines \cite{sepulveda2020robotic}. These works highlight the growing capability of robotic systems to perform complex agricultural tasks across various crops.

Research on robotic pepper harvesting has predominantly focused on protected environments where conditions are relatively controlled compared to the variability of outdoor fields. For instance, systems like HortiBot \cite{lenz2024hortibotadaptivemultiarmrobotic} and others \cite{pan2024development, lehnert2016sweet} have achieved high success rates in lab settings through environment simplification and controlled lighting. Greenhouse experiments further illustrate the benefits of a protected environment, where vertical growing systems reduce occlusion and simplify visual perception, making it easier to detect and access crops \cite{bac2017performance, lehnert2020performance, arad2020development}. In contrast, our work leverages imitation learning to conduct robotic pepper harvesting in outdoor settings. The learning-based approach enables our system to adapt and perform effectively under diverse conditions encountered in unprotected outdoor fields, showcasing a robustness that is crucial for practical agricultural applications.


The application of imitation learning (IL) has been explored across various agricultural tasks, including visual servoing for mushroom harvesting \cite{porichis2024imitation}, and tea leaf plucking \cite{9013082} using probabilistic movement primitives \cite{paraschos2013probabilistic}. For fruit harvesting, task-parameterized models have been used to automate apple picking by accommodating variations in crop positioning \cite{van2024using}, while \cite{van2024duallqr} focused on efficient grasping of oscillating apples using learning-based controllers. Additionally, support vector regression have been applied to teach robots to handle compliant food objects from demonstration \cite{misimi2018robotic}, enabling effective grasping of diverse crops. These approaches highlight the adaptability and robustness that IL can bring to agricultural robots. However, a common limitation of the aforementioned approaches are validations in controlled lab settings.
Factors like uncontrolled lighting, occlusions, physical variability in crop presentation, and external perturbations that exists in natural outdoor environments are absent in lab setups, making it difficult to apply these methods directly in the field. Additionally, the complexity of collecting high-quality demonstration data in field conditions further hinders progress. Our work aims to address these real-world challenges by demonstrating and evaluating imitation learning for agricultural robotics in the context of outdoor pepper harvesting.

\section{Preliminaries}

We first provide an overview of key methods on which our system is built, focusing on diffusion-based imitation learning and a data collection framework designed for training manipulation policies.

\subsection{Diffusion Policy \cite{chi2023diffusionpolicy}}

In this work, we employ \textit{Diffusion Policy} \cite{chi2023diffusionpolicy}, a novel imitation learning method based on conditional denoising diffusion processes \cite{ho2020denoising}, shown to be highly effective in visuomotor policy learning.
Diffusion policy offer significant advantages over existing imitation learning methods \cite{fang2019survey} attributed to its robustness in handling multimodal action distributions, capability of scaling to high-dimensional action spaces, and inherently maintaining a stable training process.
The objective of the diffusion policy is to approximate the conditional distribution \( p(\mathbf{A}_t|\mathbf{O}_t) \), which predicts actions conditioned on observations. To achieve this, the denoising diffusion probabilistic model (DDPM) learns an iterative process that refines a noisy input action into a clean, desired output. Starting from Gaussian noise, the model denoises the input over \( K \) iterations. At each step \( k \), the intermediate action \( \mathbf{A}_t^{k-1} \) is updated as:
\begin{equation}
    \mathbf{A}_t^{k-1} = \alpha (\mathbf{A}_t^k - \gamma \varepsilon_\theta (\mathbf{O}_t, \mathbf{A}_t^k, k)) + \mathcal{N} (0, \sigma^2 I)    
\end{equation}
where \( \varepsilon_\theta \) is the noise prediction network with parameters \( \theta \), and \( \mathcal{N}(0, \sigma^2 I) \) represents Gaussian noise added at each iteration. The model is trained by minimizing the mean squared error (MSE) loss:
\begin{equation}
    \mathcal{L} = \text{MSE}(\varepsilon^k, \varepsilon_\theta(\mathbf{O}_t, \mathbf{A}_t^k, k))
\end{equation}
which optimizes for the accurate prediction of the added noise \( \varepsilon^k \). Minimizing this loss effectively reduces the variational lower bound of the KL-divergence between the data distribution \( p(\mathbf{A}_t|\mathbf{O}_t) \) and the distribution of samples drawn from the DDPM, \( q(\mathbf{A}_t|\mathbf{O}_t) \).

\subsection{Universal Manipulation Interface \cite{chi2024universal}}
\label{subsec:UMI}

Collecting quality demonstration data is one of the main challenges in training a visuomotor policy through imitation learning. 
Outdoor robotic setups face additional logistical hurdles, including equipment constraints and unpredictable weather, which make consistent data collection complex and labor-intensive. To address these issues, we build upon the Universal Manipulation Interface (UMI) \cite{chi2024universal}, a data collection framework designed to train policies for complex manipulation tasks using human demonstrations. While traditional teleoperation methods are often slow, costly, and difficult to scale, and approaches relying solely on human videos suffer from the embodiment gap, UMI offers a balanced solution: a handheld, sensorized gripper that captures manipulation actions, reducing both costs and the embodiment gap.

UMI's setup involves a handheld gripper equipped with a wrist-mounted camera to collect visual and action data during human demonstrations.
The framework relies on visual-inertial SLAM (ORB-SLAM3 \cite{campos2021orb}) to generate maps for gripper localization during data collection, a process that works well in controlled, static environments typical of indoor tasks. However, directly deploying UMI in agricultural contexts presents nontrivial challenges. Firstly, agricultural tasks in outdoor fields are subject to constant disturbances like wind and contact interactions with the crop, complicating reliable localization using ORB-SLAM3. Moreover, agricultural tasks such as harvesting and pruning are typically destructive (e.g., a harvested apple cannot be reattached), necessitating frequent re-mapping for each demonstration, rendering this approach infeasible. 

In this paper, we build on UMI, but propose an alternative pose tracking method that overcomes these limitations, demonstrating how a customized handheld device can enable effective imitation learning for pepper harvesting in unprotected outdoor fields. Our approach provides a robust framework for training robotic systems to perform complex outdoor agricultural tasks. 

\section{Dataset Collection}

\subsection{Handheld Shear-Gripper}

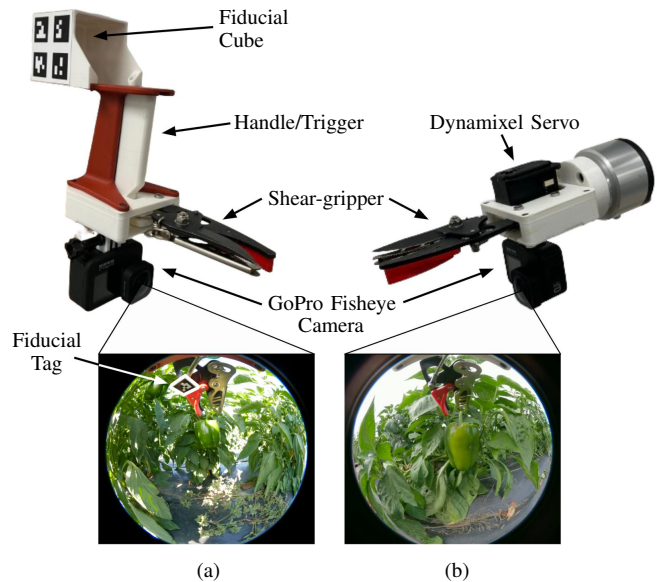
\begin{figure}[htbp]
\centering 
\includegraphics[width=\columnwidth]{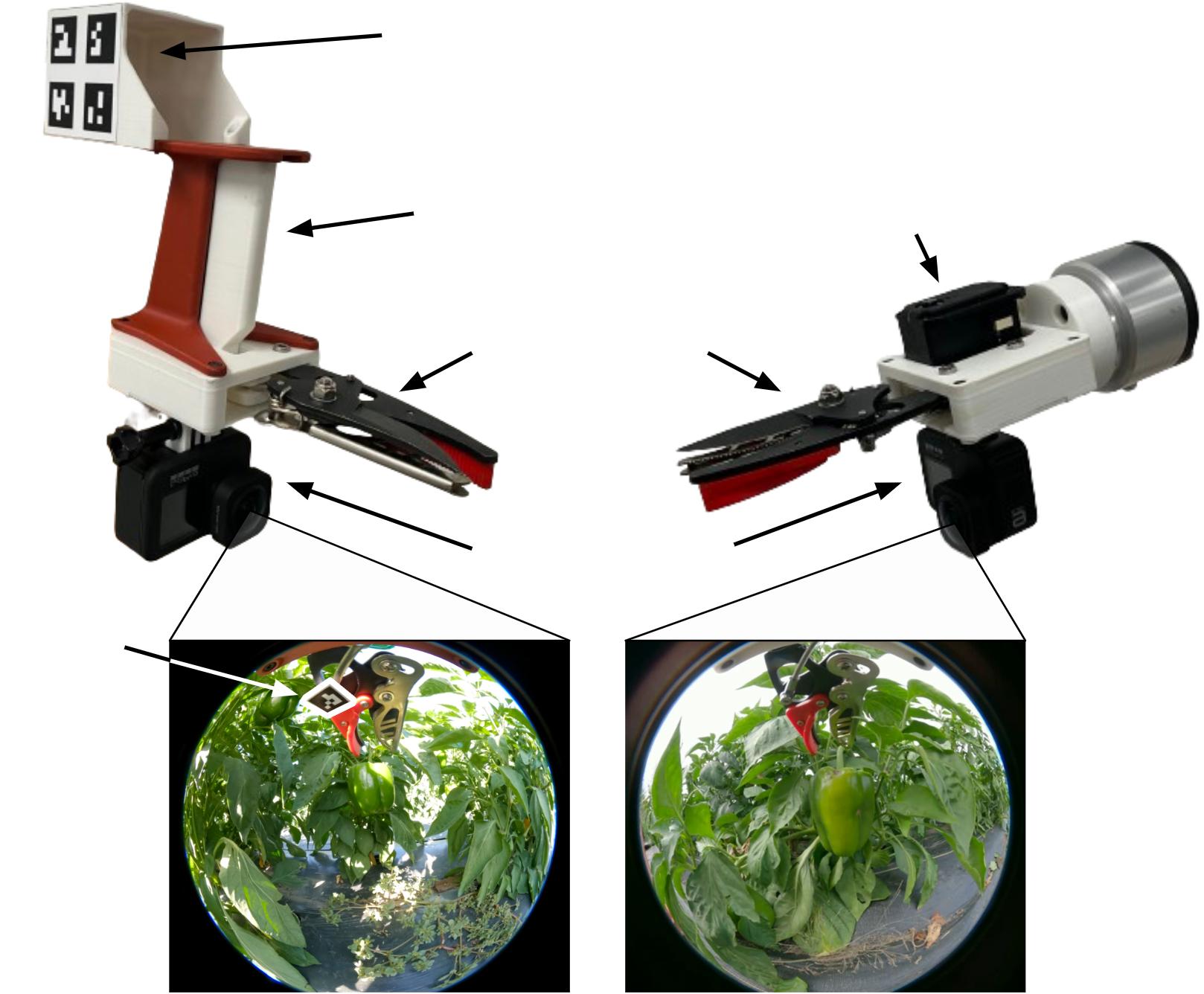}
\setlength{\unitlength}{1cm}
\begin{picture}(0,0)
\put(-1.5, 7.35){\footnotesize Fiducial}
\put(-1.5, 7.05){\footnotesize Cube}
\put(-1.3, 5.95){\footnotesize Handle/Trigger}
\put(-0.85, 4.95){\footnotesize Shear-gripper}
\put(-0.85, 3.55){\footnotesize GoPro Fisheye}
\put(-0.45, 3.25){\footnotesize Camera}
\put(1.3, 5.95){\footnotesize Dynamixel Servo}
\put(-4.25, 3.05){\footnotesize Fiducial}
\put(-4.0, 2.75){\footnotesize Tag}
\put(-1.8, 0.0){\footnotesize (a)}
\put(1.5, 0.0){\footnotesize (b)}
\end{picture}
\caption{(a) Handheld shear-gripper device used for demonstration data collection, and (b) custom robotic end-effector counterpart used for robotic deployment. The handheld device includes a fiducial cube for pose tracking via an external camera and a fiducial tag on the shear mechanism for tracking actuation. The handheld device is manually operated by the user, while the robotic end-effector is actuated by a servo motor.}
\label{fig:shear_gripper} 
\vspace{-10pt}
\end{figure}

To facilitate data collection for harvesting specialty crops in field environments, we customized the UMI framework to create a handheld device featuring a shear-gripper mechanism coupled for joint actuation (see Fig.~\ref{fig:shear_gripper}). The device was assembled using a combination of off-the-shelf components and custom 3D-printed parts, designed for ease of use and ergonomic handling. It features a handle with a trigger for the operator to manually actuate the mechanism. 
The shear-gripper is mounted below the handle, while a fiducial cube is attached on top. This arrangement is intended to reduce the physical strain of the human operator while ensuring an unobstructed view of the fiducial cube from an external camera that captures accurate positional data during demonstrations.
Additionally, a fisheye camera is mounted below the gripper ensuring that once the peduncle is cut and grasped, the harvested pepper remains clearly visible in the footage, while also capturing a fiducial tag marked on the gripper to track gripper actuation (see Fig.~\ref{fig:shear_gripper}(a)).

The handheld device closely mirrors the design of the robotic end-effector used for autonomous harvesting, with minor differences. While the robotic end-effector lacks the handle and fiducial cube found on the handheld counterpart, it maintains the same overall appearance. Actuation is provided by a Dynamixel servo motor directly connected to a linkage shaft. This similarity in design ensures minimal embodiment gap between the handheld and robotic gripper, enabling a smooth transition between human demonstrations and robot control policies.

\subsection{Robust Fiducial Cube Tracking}

\begin{figure*}[htbp]
\centering 
\includegraphics[width=0.9\textwidth]{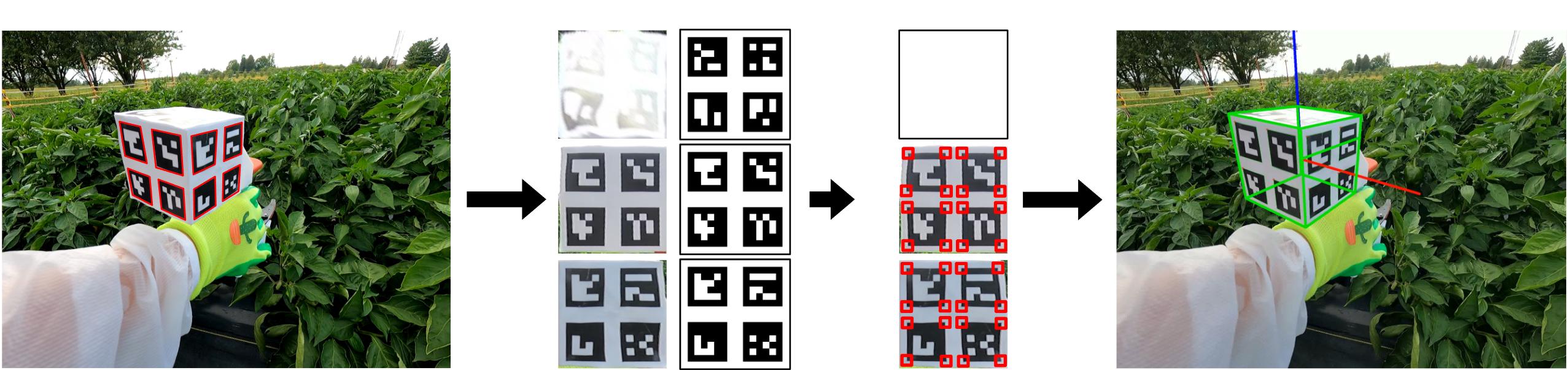}
\setlength{\unitlength}{1cm}
\begin{picture}(0,0)
\put(-15.6, 3.65){\footnotesize \CircledText{1} \textbf{Initial Pose Estimation}}
\put(-11.25, 2.7){\footnotesize \CircledText{2}}
\put(-11.45, 2.35){\footnotesize \textbf{Warp}}
\put(-11.45, 2.05){\footnotesize \textbf{Faces}}
\put(-10.35, 3.65){\footnotesize \CircledText{3} \textbf{SSIM Filter}}
\put(-8.05, 3.65){\footnotesize \CircledText{4} \textbf{Corner Refinement}}
\put(-5.6, 2.7){\footnotesize \CircledText{5}}
\put(-5.9, 2.35){\footnotesize \textbf{Unwarp}}
\put(-5.75, 2.05){\footnotesize \textbf{Faces}}
\put(-4.25, 3.65){\footnotesize \CircledText{6} \textbf{Final Pose Estimation}}
\put(-7.35, 2.83){\rotatebox[origin=c]{90}{\footnotesize {Filtered}}}
\put(-7.35, 1.63){\rotatebox[origin=c]{90}{\footnotesize {Refined}}}
\put(-7.35, 0.48){\rotatebox[origin=c]{90}{\footnotesize {Refined}}}
\end{picture}
\caption{Pipeline for robust cube pose estimation: \CircledText{1} Detect visible ArUco markers and solve the PnP problem for an initial cube pose estimate. \CircledText{2} Project the visible cube face onto a planar view based on the initial pose. \CircledText{3} Use SSIM to filter out noisy faces by comparing warped faces against expected templates. \CircledText{4} Redetect and refine marker corners on SSIM-passed faces to enhance accuracy. \CircledText{5} Map refined tag corners back to the original image space. \CircledText{6} Recompute the PnP problem using filtered, refined tag corners for precise cube pose tracking.}
\label{fig:robust_filter} 
\vspace{-15pt}
\end{figure*}
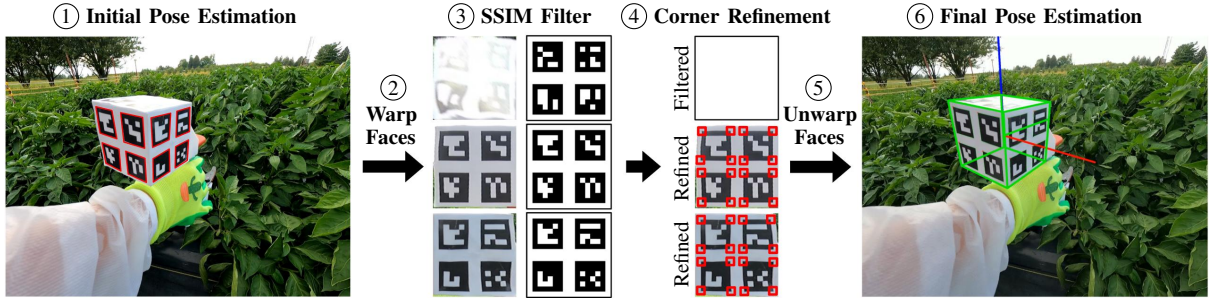

As discussed in Sec.~\ref{subsec:UMI}, UMI's visual-inertial SLAM-based data collection method is impractical for agricultural environments due to constant field disturbances and the need for frequent remapping. To address this problem, we adopted an approach involving the tracking of a fiducial cube attached to the handheld device. This solution requires an external camera providing a third-eye view of the cube during data collection as illustrated in Fig.~\ref{fig:robust_filter}, but eliminates the need for the inadequate SLAM-based localization. Note that the additional camera is necessary only during the data collection phase and is not required during the actual deployment of the robot.

The cube is designed with 2$\times$2 ArUco markers on four of its faces. A naive implementation of cube tracking via Perspective-n-Point (PnP) was found to produce noisy and unstable gripper pose estimations, rendering the data unsuitable for precise tracking. To improve robustness, we introduce a post-processing method to ensure consistent and accurate pose estimation across frames. The process consists of the following steps (refer to Fig.~\ref{fig:robust_filter}):
\begin{enumerate} 
    \item \textbf{Initial Pose Estimation}: Detect all visible fiducial markers and solve the PnP-problem, obtaining an initial estimate of the cube's pose. 
    \item \textbf{Warp Cube Faces}: Compute the corners of the visible cube faces using the initial pose and project them onto a planar view. 
    \item \textbf{Filtering via SSIM}: Compare the warped face with the expected template using the Structural Similarity Index Measure (SSIM) to filter out noisy faces based on a set threshold. 
    \item \textbf{Corner Refinement}: For faces passing the SSIM filter, redetect fiducial markers on the planar view to obtain refined tag corner pixel coordinates. 
    \item \textbf{Unwarp Cube Faces}: Map the refined tag coordinates back to the original image space to obtain precise pixel positions. 
    \item \textbf{Final Pose Estimation}: Recompute the PnP-problem using the filtered, refined tag corners to obtain accurate cube pose estimation. 
\end{enumerate}

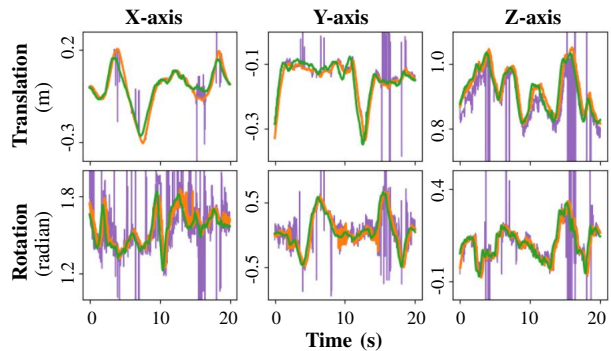
\begin{figure}[htbp]
\vspace{4pt}
\centering 
\includegraphics[width=0.9\columnwidth]{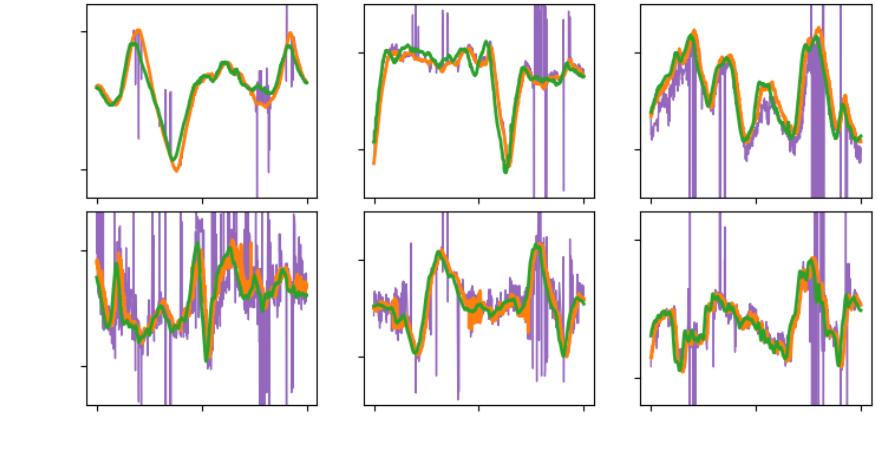}
\setlength{\unitlength}{1cm}
\begin{picture}(0,0)
\put(-6.7, 4.05){\footnotesize \textbf{X-axis}}
\put(-4.2, 4.05){\footnotesize \textbf{Y-axis}}
\put(-1.65, 4.05){\footnotesize \textbf{Z-axis}}
\put(-8.2, 3.0){\rotatebox[origin=c]{90}{\footnotesize \textbf{Translation}}}
\put(-7.9, 3.0){\rotatebox[origin=c]{90}{\footnotesize (m)}}
\put(-8.2, 1.05){\rotatebox[origin=c]{90}{\footnotesize \textbf{Rotation}}}
\put(-7.9, 1.05){\rotatebox[origin=c]{90}{\footnotesize (radian)}}

\put(-7.2, 0.05){\scriptsize 0}
\put(-6.35, 0.05){\scriptsize 10}
\put(-5.45, 0.05){\scriptsize 20}

\put(-4.75, 0.05){\scriptsize 0}
\put(-3.9, 0.05){\scriptsize 10}
\put(-2.95, 0.05){\scriptsize 20}

\put(-2.3, 0.05){\scriptsize 0}
\put(-1.45, 0.05){\scriptsize 10}
\put(-0.55, 0.05){\scriptsize 20}

\put(-7.55, 3.65){\rotatebox[origin=c]{90}{\scriptsize 0.2}}
\put(-7.55, 2.45){\rotatebox[origin=c]{90}{\scriptsize -0.3}}
\put(-5.1, 3.45){\rotatebox[origin=c]{90}{\scriptsize -0.1}}
\put(-5.1, 2.6){\rotatebox[origin=c]{90}{\scriptsize -0.3}}
\put(-2.63, 3.45){\rotatebox[origin=c]{90}{\scriptsize 1.0}}
\put(-2.63, 2.55){\rotatebox[origin=c]{90}{\scriptsize 0.8}}

\put(-7.55, 1.65){\rotatebox[origin=c]{90}{\scriptsize 1.8}}
\put(-7.55, 0.65){\rotatebox[origin=c]{90}{\scriptsize 1.2}}
\put(-5.1, 1.6){\rotatebox[origin=c]{90}{\scriptsize 0.5}}
\put(-5.1, 0.7){\rotatebox[origin=c]{90}{\scriptsize -0.5}}
\put(-2.63, 1.75){\rotatebox[origin=c]{90}{\scriptsize 0.4}}
\put(-2.63, 0.55){\rotatebox[origin=c]{90}{\scriptsize -0.1}}

\put(-4.3, -0.25){\footnotesize \textbf{Time (s)}}
\end{picture}
\vspace{5pt}
\caption{The plot shows the 6-DOF pose of the fiducial cube captured during a single demonstration in a static environment. Our filter/refine method ({\color{orange}orange line}) significantly reduces noise when compared to the direct PnP method ({\color{violet}purple line}). The tracking result from ORB-SLAM3 is also plotted in {\color{ForestGreen}green}.}
\label{fig:filter_results} 
\vspace{-10pt}
\end{figure}

We quantitatively evaluated our method by using the trajectory from ORB-SLAM3 as the baseline, as illustrated in Fig.~\ref{fig:filter_results}. The trajectory was derived in a static environment to avoid the known issue of ORB-SLAM3 breaking in dynamic scenes. Without refinement (purple line), the pose estimates were noisy, leading to mean squared errors (MSE) of 0.276 for position and 0.358 for rotation. Applying our method (orange line) significantly reduced noise, achieving MSEs of 0.003 and 0.018, respectively, yielding a smoother trajectory that supports reliable data collection for robust policy training. This approach bypasses the limitations of SLAM-based localization, enabling data collection in dynamic agricultural settings.

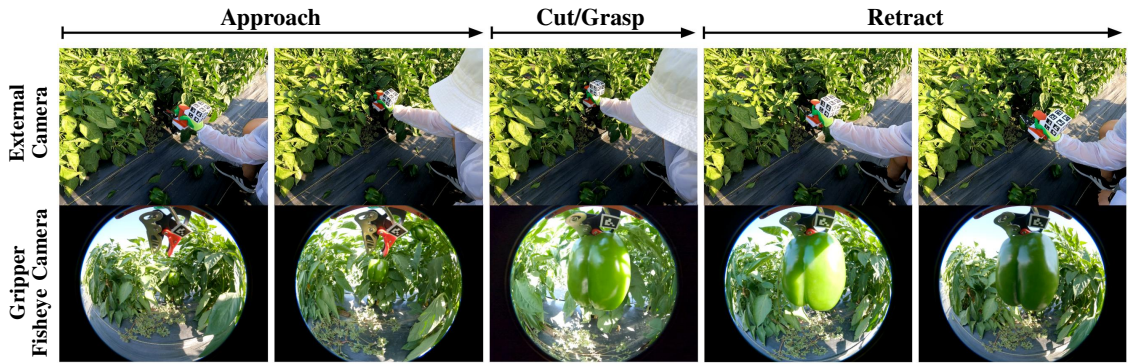
\begin{figure*}[t]
\centering 
\includegraphics[width=0.8\textwidth]{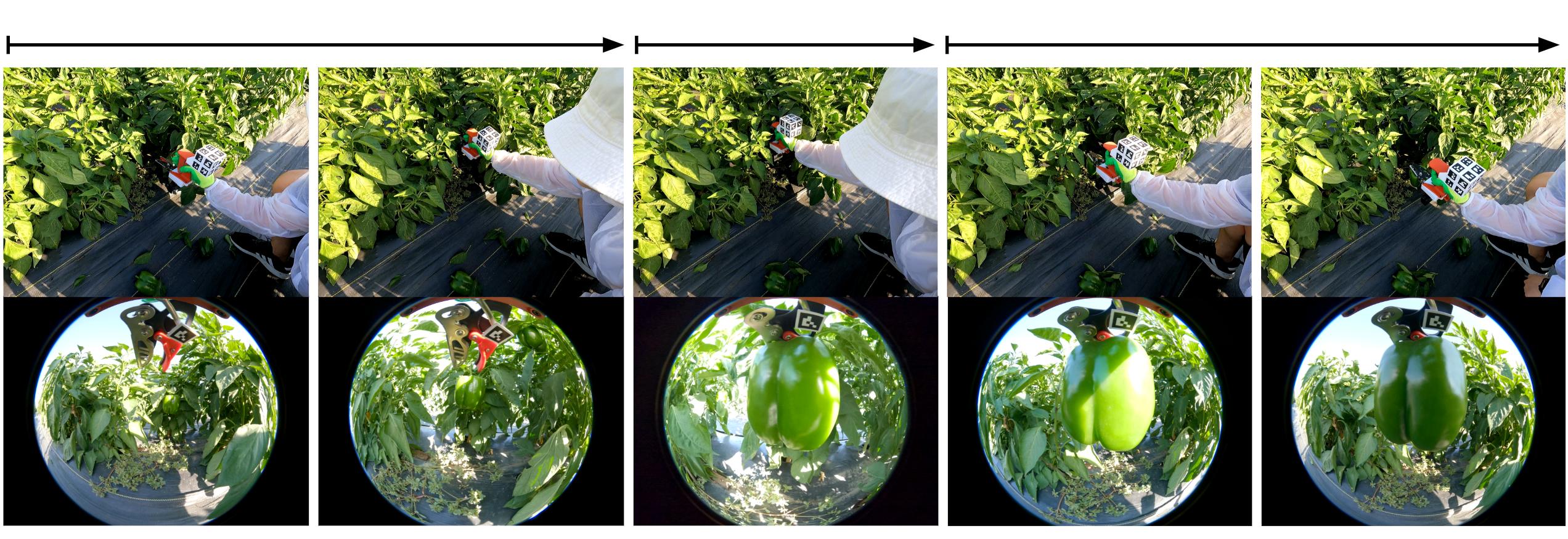}
\setlength{\unitlength}{1cm}
\begin{picture}(0,0)
\put(-12.3, 4.6){\small \textbf{Approach}}
\put(-8.1, 4.6){\small \textbf{Cut/Grasp}}
\put(-3.7, 4.6){\small \textbf{Retract}}
\put(-15.1, 3.2){\rotatebox[origin=c]{90}{\footnotesize \textbf{External}}}
\put(-14.8, 3.2){\rotatebox[origin=c]{90}{\footnotesize \textbf{Camera}}}
\put(-15.1, 1.1){\rotatebox[origin=c]{90}{\footnotesize \textbf{Gripper}}}
\put(-14.8, 1.1){\rotatebox[origin=c]{90}{\footnotesize \textbf{Fisheye Camera}}}
\end{picture}
\vspace{-5pt}
\caption{A harvesting demonstration begins with the operator approaching the pepper with the handheld shear-gripper device, followed by cutting and grasping the pepper’s peduncle, and finally retracting from the pepper plant. The top row displays the external camera's viewpoint capturing the fiducial cube for tracking, while the bottom row displays the viewpoint from the gripper's fisheye camera.}
\label{fig:demonstration} 
\vspace{-15pt}
\end{figure*}

\subsection{Pepper Harvesting Data Collection}

To collect demonstration data for training a pepper harvesting visuomotor policy, we conducted a field trip to the Iowa State University Horticulture Research Station, where 300 human demonstrations were recorded over a 2-day period. 

The demonstration process is illustrated in Fig.~\ref{fig:demonstration}. Each episode begins with the human operator positioning the handheld device such that the target pepper is visible within the field of view of the gripper camera. The external camera  is also setup on a tripod ensuring that the fiducial cube remains within the external camera's field of view. While both cameras are recording, the operator approaches the pepper, maneuvering the device to align the peduncle of the target pepper between the jaws of the shear-gripper. Once in position, the operator actuates the shear-gripper to cut and grasp the peduncle of the pepper. The device then retracts from the pepper plant, completing the harvesting demonstration.

The demonstrations composed from the pair of videos can be used to train a visuomotor policy that takes as input a sequence of observations (RGB images, gripper pose, gripper actuation) and outputs a sequence of actions (gripper pose and gripper actuation). Given the scarcity and value of agricultural datasets, we are releasing this dataset to the research community to facilitate further research and development in robotic agricultural systems.

\section{System Deployment}

\subsection{Outdoor Pepper Field}

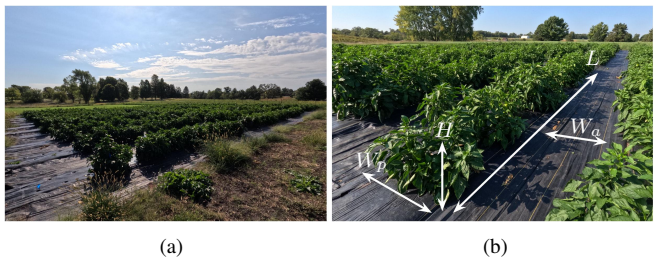
\begin{figure}[htbp]
\vspace{8pt}
\centering 
\includegraphics[width=\columnwidth]{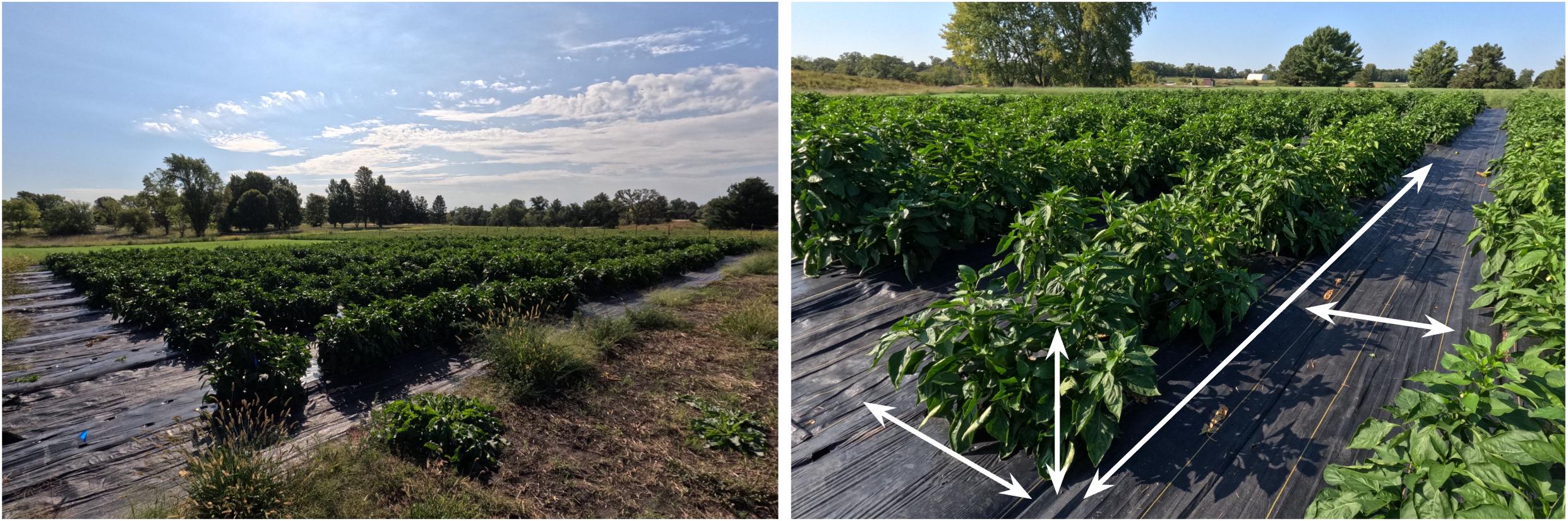}
\setlength{\unitlength}{1cm}
\begin{picture}(0,0)
\put(-2.25, 0.0){\footnotesize (a)}
\put(2.05, 0.0){\footnotesize (b)}
\put(0.35, 1.15){\footnotesize {\color{white}$W_p$}}
\put(1.4, 1.55){\footnotesize {\color{white}$H$}}
\put(3.4, 2.5){\footnotesize {\color{white}$L$}}
\put(3.2, 1.6){\footnotesize {\color{white}$W_a$}}
\end{picture}
\caption{We conducted data collection and experiments at an unprotected outdoor pepper field at the Iowa State University Horticultural Research Station. (a) The pepper field was composed of 10 rows, each row divided into 4 plots of pepper plants. (b) Each plot measured approximately 1.0m in width ($W_p$), 6.6m in length ($L$), and 0.7m in height ($H$), while the aisles between rows ($W_a$) were about 1.0m wide.}
\label{fig:pepper_field} 
\vspace{-10pt}
\end{figure}

The experimental evaluation consisted of 221 trials conducted over a 3-day period at an outdoor pepper field shown in Fig.~\ref{fig:pepper_field}(a). The pepper field was composed of 10 rows, with each row containing 4 plots of pepper plants. Each plot (see Fig.~\ref{fig:pepper_field}(b)) was approximately 1.0m in width, 6.6m in length, and 0.7m in height, with the width of aisles in between rows measuring around 1.0m. On average, each plot yielded approximately 30 peppers. For our study, Iowa State University generously provided access to 5 rows. We designated 2.5 rows (equivalent to 10 plots) for collecting demonstration data and reserved the remaining 2.5 rows (10 plots) for testing the robotic system. This setup allowed us to collect comprehensive data and thoroughly evaluate our approach in a commercial agricultural environment.

\subsection{Hardware Setup}

\begin{figure}[htbp]
\centering 
\includegraphics[width=0.65\columnwidth]{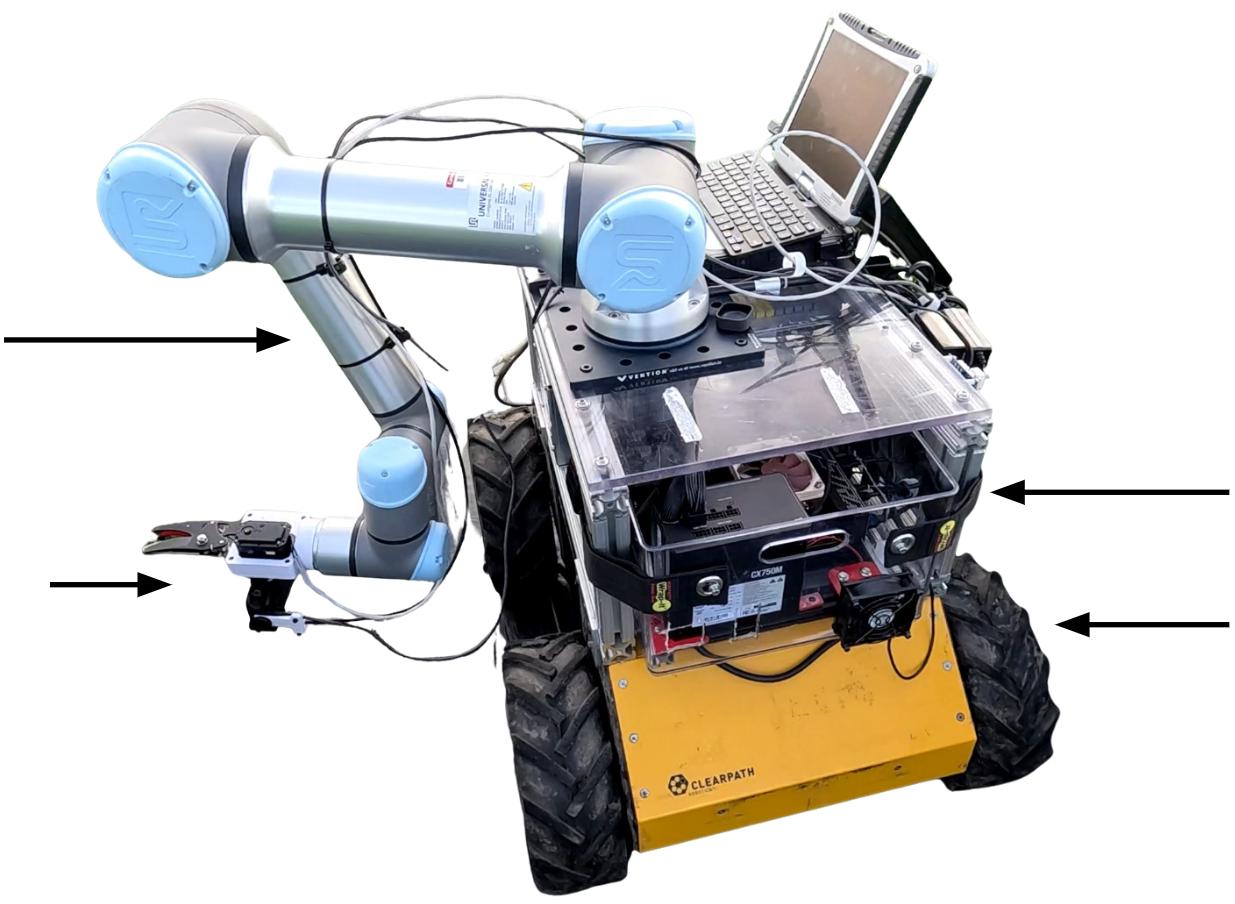}
\setlength{\unitlength}{1cm}
\begin{picture}(0,0)
\put(-6.85, 2.65){\footnotesize UR5e}
\put(-7.2, 2.35){\footnotesize Robot Arm}
\put(-7.2, 1.55){\footnotesize Shear-gripper}
\put(-7.15, 1.25){\footnotesize End-effector}
\put(-0.2, 1.95){\footnotesize On-board}
\put(-0.15, 1.65){\footnotesize Computer}
\put(-0.2, 1.2){\footnotesize Husky UGV}
\end{picture}
\vspace{-5pt}
\caption{Our hardware setup for autonomous pepper harvesting, featuring a custom shear-gripper end-effector attached to the UR5e robot arm. The robot arm is mounted on the Husky UGV that serves as a mobile platform.}
\label{fig:robot} 
\vspace{-15pt}
\end{figure}
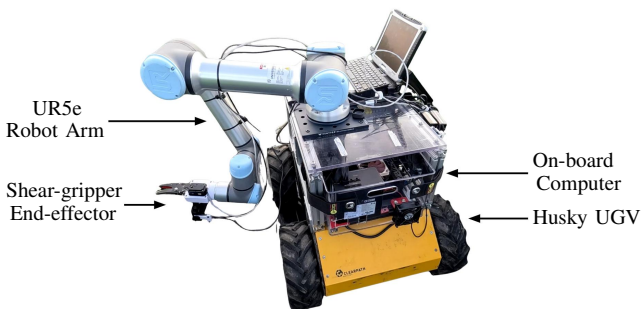

Our hardware setup for the deployment of automated pepper harvesting (see Fig.~\ref{fig:robot}) features a UR5e robot arm mounted on a Husky ground robot serving as a mobile platform. The Husky houses the UR5e control box and an on-board computer with an Intel Core i9-13900K processor and a GeForce RTX 4070 12GB GPU. Attached to the UR5e arm is the shear-gripper end-effector (Fig.~\ref{fig:shear_gripper}(b)), which performs the cutting and grasping of peppers during automated harvesting. Unlike the data collection setup, the system no longer needs the tripod mounted external camera; only the camera mounted on the end-effector is required for deployment.

\subsection{Our Software}

Our software for robotic pepper harvesting includes a diffusion policy trained on the dataset of 300 pepper harvesting demonstrations. The visuomotor policy controls the robot arm to approach and grasp a pepper, but additional functionality is required to complete the placement of the harvested pepper. To achieve this, we implemented a grasp detector with a simple convolutional neural network (CNN) architecture. The network comprises two convolutional layers and a fully connected output layer, taking RGB images as input and outputting binary classification to indicate whether a pepper has been successfully grasped.

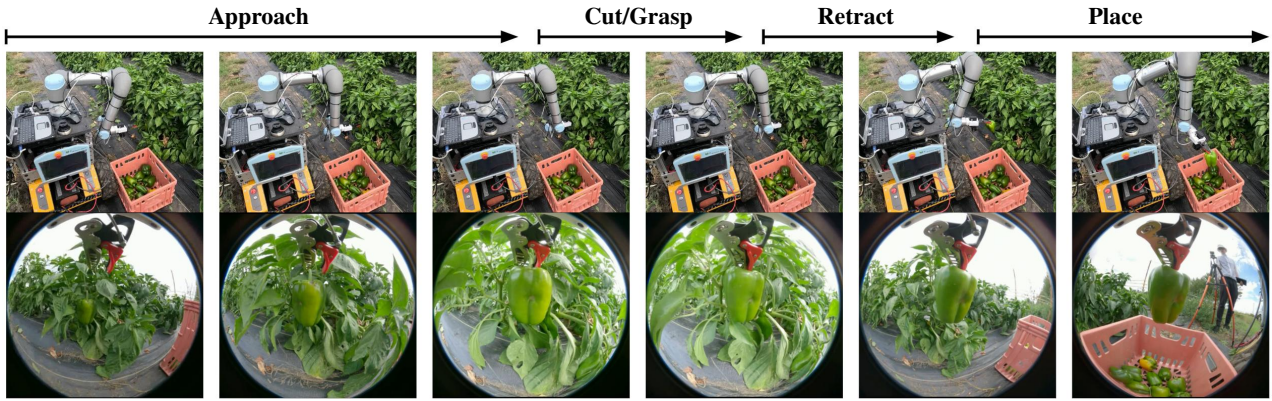
\begin{figure*}[htbp]
\centering 
\includegraphics[width=0.95\textwidth]{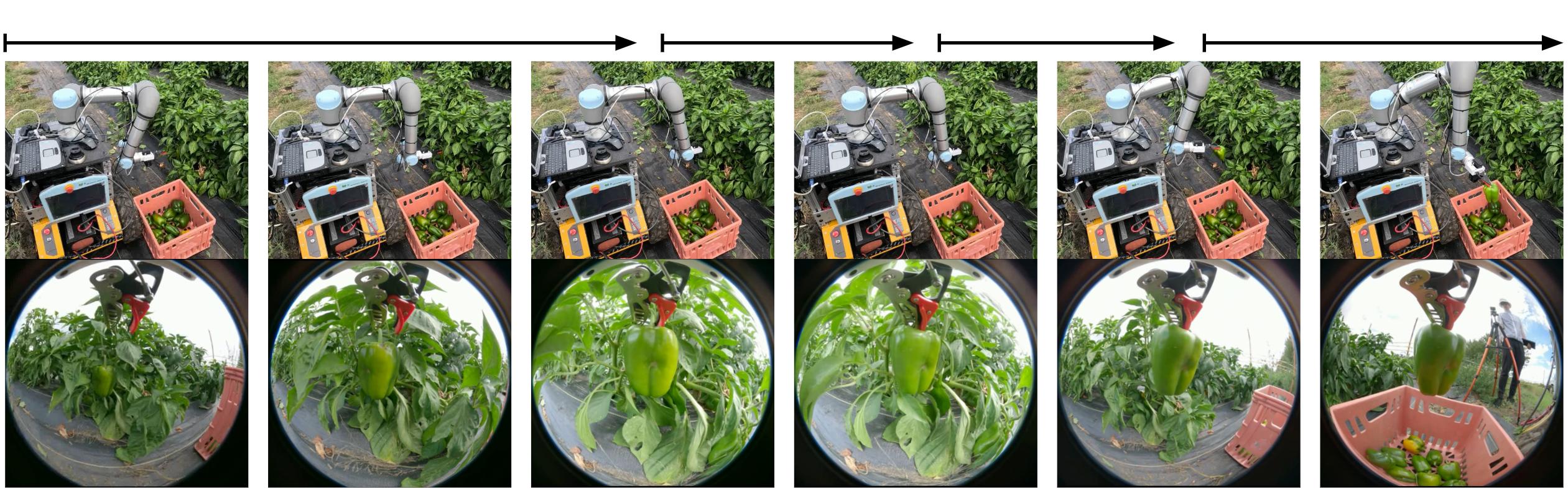}
\setlength{\unitlength}{1cm}
\begin{picture}(0,0)
\put(-14.4, 5.0){\small \textbf{Approach}}
\put(-9.4, 5.0){\small \textbf{Cut/Grasp}}
\put(-6.3, 5.0){\small \textbf{Retract}}
\put(-2.7, 5.0){\small \textbf{Place}}
\end{picture}
\caption{Autonomous pepper harvesting by the robot. The approach, cut, grasp, and retraction actions are guided by a trained visuomotor policy learned from demonstrations, while the placement is executed in an open-loop manner upon detecting a successful grasp. The top row images are included for illustration purposes only; the policy operates without external camera observations during deployment.}
\label{fig:deployment} 
\vspace{-15pt}
\end{figure*}

Each pepper harvesting trial consists of two phases: a manual setup phase and an autonomous harvesting phase. In the manual setup phase, the Husky ground robot is manually positioned next to the target pepper plant. The end-effector is then adjusted via joystick to ensure the target pepper is within the gripper camera’s field of view, preparing the system for autonomous operation. During the autonomous harvesting phase, control of the robot arm is transferred to the visuomotor policy, which directs the end-effector to approach, cut, and grasp the pepper’s peduncle before retracting as illustrated in Fig.~\ref{fig:deployment}. Once a successful grasp is detected by the grasp detector, an open-loop motion controller guides the arm to place the harvested pepper into a crate positioned beside the Husky UGV.


Automating the manual setup phase was beyond the scope of this work, as our objective was to evaluate the manipulation phase which requires precise contact interaction between the robot and the crop—a challenge we approached through imitation learning. Automation of the setup phase remains an area for future development.


\section{Experiment Results}

\subsection{Task Difficulty Categorization}

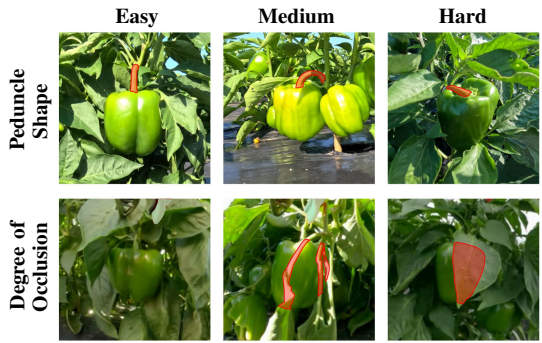
\begin{figure}[htbp]
\vspace{15pt}
\centering 
\includegraphics[width=0.8\columnwidth]{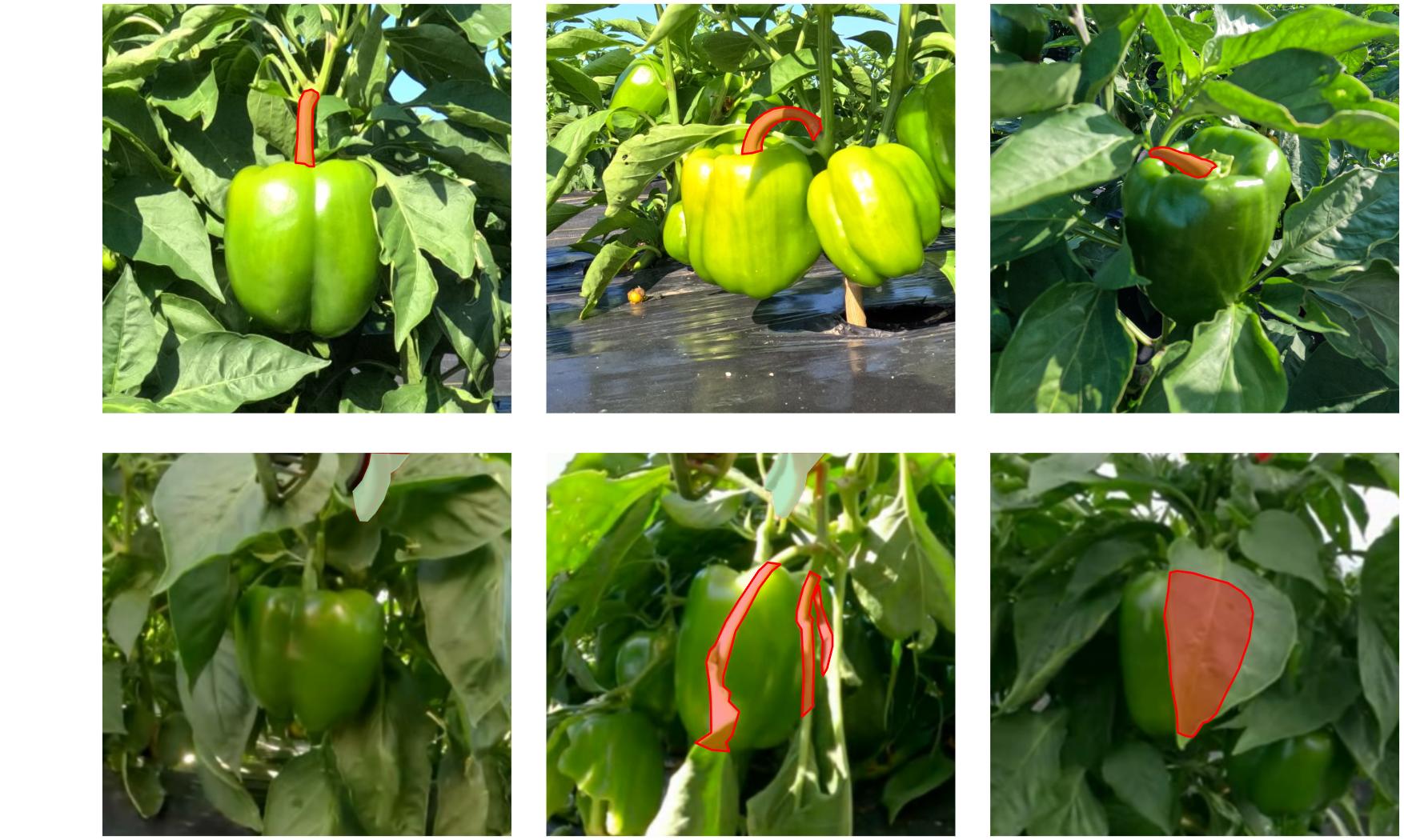}
\setlength{\unitlength}{1cm}
\begin{picture}(0,0)
\put(-5.9, 4.25){\footnotesize \textbf{Easy}}
\put(-4.0, 4.25){\footnotesize \textbf{Medium}}
\put(-1.6, 4.25){\footnotesize \textbf{Hard}}
\put(-7.3, 3.0){\rotatebox[origin=c]{90}{\footnotesize \textbf{Peduncle}}}
\put(-7.0, 3.0){\rotatebox[origin=c]{90}{\footnotesize \textbf{Shape}}}
\put(-7.3, 0.95){\rotatebox[origin=c]{90}{\footnotesize \textbf{Degree of}}}
\put(-7.0, 0.9){\rotatebox[origin=c]{90}{\footnotesize \textbf{Occlusion}}}
\end{picture}
\caption{Target peppers are categorized into easy, medium, and hard difficulty levels based on peduncle shape and degree of occlusion. Peduncle shapes (top row) and occlusion regions (bottom row) are highlighted in red.}\label{fig:difficulty_level} 
\vspace{-15pt}
\end{figure}

To assess the performance of the visuomotor policy, we categorized the difficulty of each trial based on two key features: peduncle shape and degree of occlusion of the target pepper. Each feature was further divided into three difficulty levels (refer to Fig.~\ref{fig:difficulty_level}):

\subsubsection{Peduncle Shape}
\begin{itemize}
    \item \textbf{Easy:} Ideal peduncles that are oriented vertically, providing ample space for the gripper fingers to easily grasp and cut.
    \item \textbf{Medium:} More challenging peduncles that are slanted or curved. This orientation narrows the gap between the peduncle and the pepper, obstructing the gripper and increasing the difficulty of the task.
    \item \textbf{Hard:} Peduncles that are impeded by the stem or lie flat against the pepper, making it difficult to reach and grasp them effectively with the current gripper design.
\end{itemize}
\subsubsection{Degree of Occlusion}
\begin{itemize}
    \item \textbf{Easy:} Minimal occlusion, where the pepper and its peduncle are almost fully visible, allowing for straightforward detection and approach.
    \item \textbf{Medium:} Partially occluded peppers, with less than half of the pepper obscured by leaves or other parts of the plant. 
    \item \textbf{Hard:} Major occlusion, where more than half of the pepper is hidden, posing significant challenges for detection and harvesting.
\end{itemize}

\subsection{Evaluation Results}

The visuomotor policy and grasp detector were evaluated independently to assess their contributions to the overall pepper harvesting process.

\subsubsection{Harvesting Policy Evaluation}
The visuomotor policy’s performance was evaluated based on its ability to successfully approach, cut, and grasp a pepper, defining success as the completion of these tasks without considering the open-loop placement phase which relies on the grasp detector’s accuracy. Quantitative results are summarized in Table~\ref{table:results}, detailing success rates across varying difficulty levels. Overall, the policy achieved a total success rate of 28.95\% across 221 trials with an average cycle time of 31.7s. The success rates decreased as task difficulty increased, reflecting the impact of environmental complexity on policy performance.

\begin{table}[h]
\vspace{-5pt}
\captionsetup{font=small} 
\caption{Evaluation results of harvesting visuomotor policy }
\vspace{-5pt}
\centering
\setlength{\tabcolsep}{24pt} 
\renewcommand{\arraystretch}{1.3} 
\begin{tabularx}{\columnwidth}{@{}ccc@{}} 
\hline\hline
\textbf{Difficulty} & \textbf{Success Rate} & \textbf{Cycle Time (s)} \\ 
\hline 
Easy   & 42.57\% (43/101) & 28.94 \\
Medium & 20.21\% (19/93)   & 40.08 \\
Hard   & 7.31\% (2/27)    & 25.71  \\
\hdashline
\rule{0pt}{2.5ex}
\textbf{Total}  & \textbf{28.95\%} \textbf{(64/221)}  & \textbf{31.71} \\
\hline\hline
\end{tabularx}
\label{table:results} 
\vspace{-5pt}
\end{table}

\subsubsection{Grasp Detector Evaluation}

\begin{figure}[htbp]
\vspace{-5pt}
\centering 
\includegraphics[width=\columnwidth]{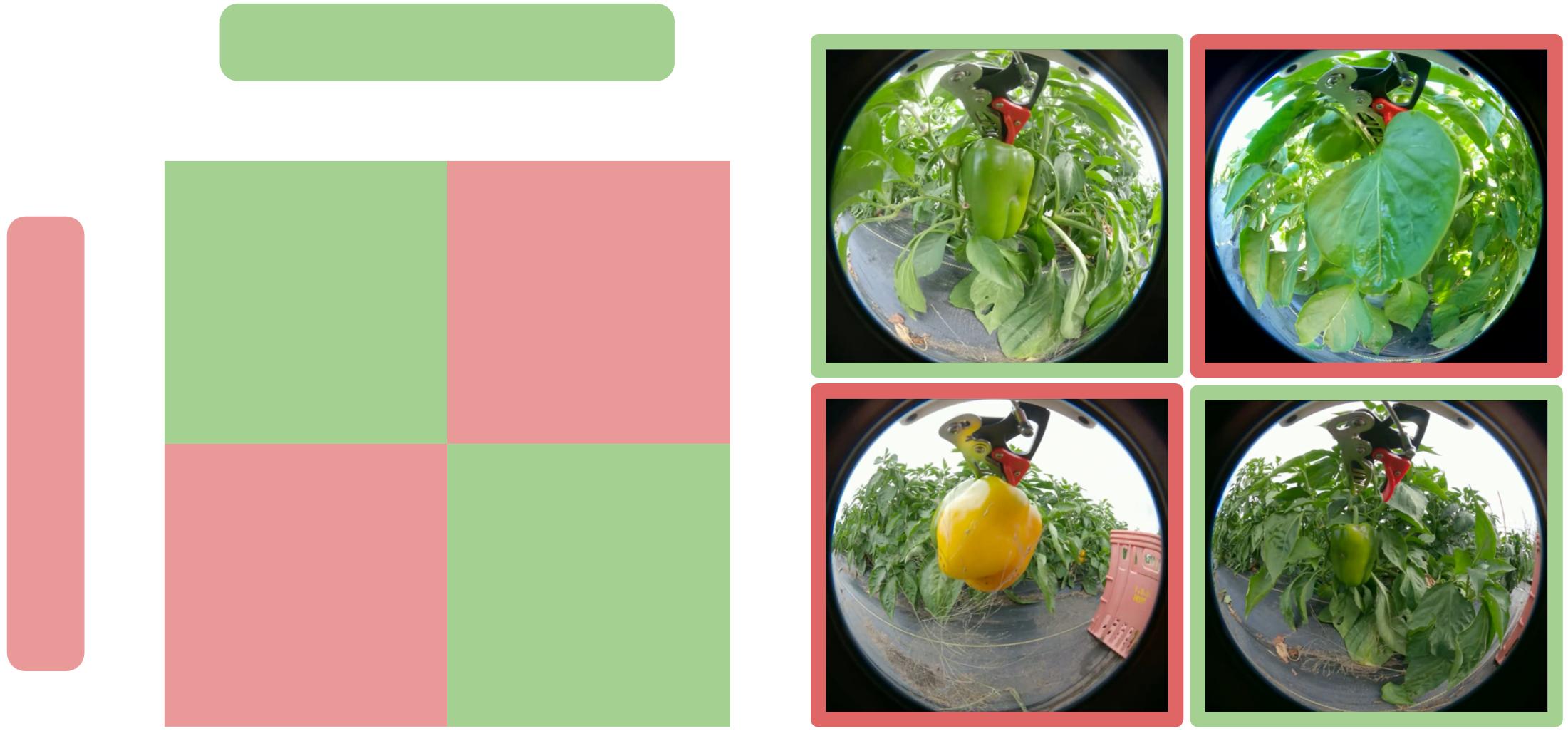}
\setlength{\unitlength}{1cm}
\begin{picture}(0,0)
\put(-2.55, 4.13){\footnotesize \fontfamily{phv}\selectfont \textbf{True Grasp}}
\put(-3.0, 3.65){\scriptsize \fontfamily{phv}\selectfont Positive}
\put(-1.5, 3.65){\scriptsize \fontfamily{phv}\selectfont Negative}
\put(-3.7, 2.7){\rotatebox[origin=c]{90}{\scriptsize \fontfamily{phv}\selectfont Positive}}
\put(-3.7, 1.2){\rotatebox[origin=c]{90}{\scriptsize \fontfamily{phv}\selectfont Negative}}
\put(-4.12, 1.95){\rotatebox[origin=c]{90}{\footnotesize \fontfamily{phv}\selectfont \textbf{Predicted Grasp}}}
\put(-2.75, 2.85){\small \fontfamily{phv}\selectfont \textbf{TP}}
\put(-2.73, 2.55){\small \fontfamily{phv}\selectfont 54}
\put(-1.22, 2.85){\small \fontfamily{phv}\selectfont \textbf{FP}}
\put(-1.20, 2.55){\small \fontfamily{phv}\selectfont 22}
\put(-2.75, 1.25){\small \fontfamily{phv}\selectfont \textbf{FN}}
\put(-2.73, 0.95){\small \fontfamily{phv}\selectfont 15}
\put(-1.22, 1.25){\small \fontfamily{phv}\selectfont \textbf{TN}}
\put(-1.27, 0.95){\small \fontfamily{phv}\selectfont 123}
\put(-2.0, 0.05){\footnotesize (a)}
\put(2.1, 0.05){\footnotesize (b)}
\end{picture}
\vspace{-5pt}
\caption{(a) Confusion matrix summarizing the grasp detector's performance. (b) Visual examples from each quadrant: in the FP quadrant, the detector incorrectly identifies a leaf as a grasped pepper, while in the FN quadrant, it fails to detect a grasped yellow pepper unseen during training.}
\label{fig:grasp_detector} 
\vspace{-15pt}
\end{figure}
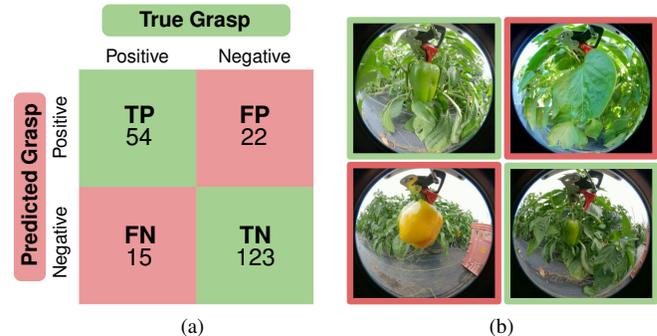

The grasp detector’s performance was evaluated based on its ability to accurately signal the transition from visuomotor policy control to the open-loop placement phase upon detecting a successfully grasped pepper. The confusion matrix displayed in Fig.~\ref{fig:grasp_detector}(a) provides a quantitative summary of the detector’s performance. Accuracy, at 0.83, indicates the success rate in correct detections. Precision, measuring true positive detections among all grasp detections, is 0.71, while recall, indicating true positive detections among actual successful grasps, is 0.78. The F1-score stands at 0.74, reflecting the detector's overall consistency in grasp recognition.

\subsection{Comparison with Prior Works}
Previous studies on robotic pepper harvesting conducted in greenhouse setups \cite{bac2017performance, lehnert2020performance, arad2020development} often tested on modified peppers, where crops were selectively pruned or repositioned to facilitate easier harvesting by robots, thereby improving success rates. This approach made target peppers more accessible and less obscured, allowing systems to operate under controlled conditions. Additionally, these studies utilized vertical growing structures, which generally improve crop accessibility by reducing occlusions and simplifying reachability. In contrast, unmodified peppers are left in their natural, unaltered state, presenting the typical challenges of real-world agricultural environments, such as varying orientations, occlusions, and natural growth patterns. To align our evaluation with these conditions, we associate the easy category with modified peppers, while the combined medium and hard categories represent unmodified peppers. As reported in Table~\ref{table:prior_work_comparison}, our success rates are comparable to those reported in previous works, despite the added complexity of operating in unprotected outdoor fields.

\begin{table}[h]
\vspace{-5pt}
\captionsetup{font=small} 
\caption{Harvesting Performance versus Prior Works}
\vspace{-5pt}
\centering
\setlength{\tabcolsep}{3.5pt} 
\renewcommand{\arraystretch}{1.3} 
\begin{tabularx}{\columnwidth}{@{}ccccc@{}} 
\hline\hline
\textbf{Baselines} & \textbf{Setting} & \textbf{Unmodified} & \textbf{Modified} & \textbf{Cycle Time (s)} \\ 
\hline 
\cite{bac2017performance} & Greenhouse & 4\% (4/90) & 29\% (25/86)  & 105.8 \\
\cite{lehnert2020performance} & Glasshouse & 25\% & 31\% (53/170)  & 36.9 \\
\cite{arad2020development} & Greenhouse & 18\% (29/159) & 49\% (51/104)  & 24.0 \\
Ours & Outdoor field & 18\% (21/120) & 43\% (43/101)  & 31.7 \\
\hline\hline
\end{tabularx}
\captionsetup{justification=centering}
\begin{minipage}{\columnwidth}
\vspace{1mm}
\footnotesize 
\begin{itemize}
    \item \cite{bac2017performance} \textit{Success rates were adjusted by summing the outcomes for both fin-ray and lip-type end-effectors.}
    \item \cite{lehnert2020performance} \textit{Success rates were calculated based on the reported harvest success rate over the average number of attempts.} 
    \item \cite{arad2020development} \textit{Success rates for the double-row setup are reported for comparative conditions with our outdoor field setting.}
\end{itemize}
\end{minipage}
\label{table:prior_work_comparison}
\vspace{-10pt}
\end{table}


\section{Discussion}

The results of our experiments highlight both the successes and limitations of our autonomous pepper harvesting system, providing valuable insights for future improvements.

\subsection{Successes}

\begin{figure}[htbp]
\vspace{-5pt}
\centering 
\includegraphics[width=\columnwidth]{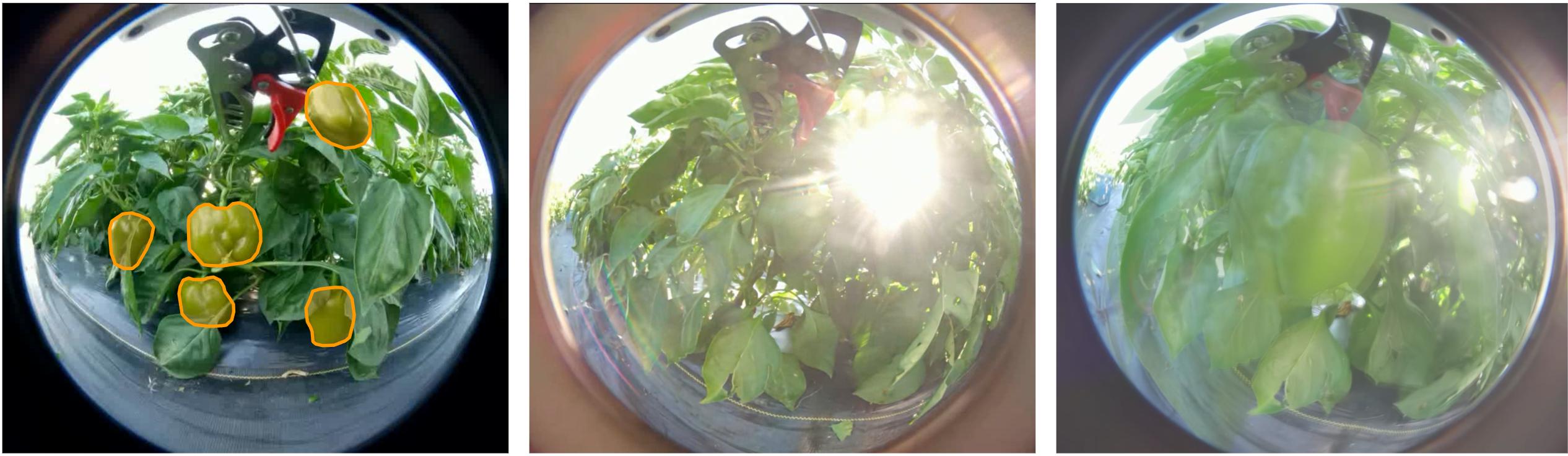}
\setlength{\unitlength}{1cm}
\begin{picture}(0,0)
\put(-3.0, 0.1){\footnotesize (a)}
\put(-0.1, 0.1){\footnotesize (b)}
\put(2.8, 0.1){\footnotesize (c)}
\end{picture}
\vspace{-5pt}
\caption{(a) Various scenarios are encountered, such as instances where five peppers (highlighted yellow) appear within the camera’s field of view. (b) Exposure to direct sunlight introduces significant lighting variability in the camera feed. (c) Overlaying three consecutive, partially transparent images sampled at 10Hz highlights the dynamic nature of the environment.}
\label{fig:successes} 
\vspace{-15pt}
\end{figure}
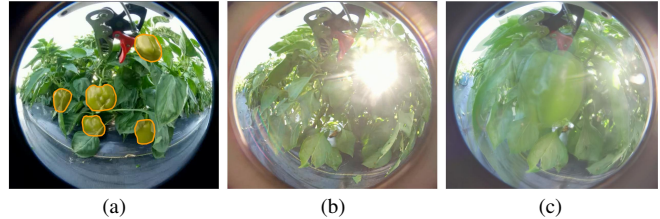

One of the key strengths of our approach lies in its generalizability, achieved through our versatile data collection system. By enabling data collection from multiple locations and diverse environmental conditions, we were able to train a policy that effectively adapts to new and unseen settings (e.g. Fig.~\ref{fig:successes}(a)). This variety in data, encompassing diverse pepper shapes and field conditions, enabled the system to generalize well across different locations and scenarios, enhancing overall robustness and adaptability. 

Our system also demonstrated notable robustness in handling dynamic scenes and fluctuating lighting conditions, typical of outdoor agricultural settings as illustrated in Fig.~\ref{fig:successes}. This resilience can be attributed to the advantages of the diffusion policy combined with diverse training data. Beyond handling variations in lighting and scene composition, the policy also exhibited reactive behaviors, such as the emergence of recovery strategies. For example, in cases where a pepper was not successfully retained in the gripper, the policy autonomously reattempted the harvesting process (a behavior that was not part of any demonstration), illustrating a self-corrective ability that supports continuous autonomous operation in unpredictable environments.

\subsection{Failure Cases} 

Several failure cases highlighted areas for potential improvement. The most frequent issue involved the gripper’s inability to position the peduncle accurately between its fingers, resulting in missed cuts. This primarily occurred because the image encoder in the diffusion policy was not explicitly trained to identify the peduncle; instead, it appeared to focus primarily on the pepper fruit. Consequently, the policy’s actions tended to position the gripper relative to the fruit rather than accurately targeting the peduncle, which is crucial for a successful cut.

Another common failure involved the occasional dropping of peppers after a successful cut. This issue arose from the fact that many human demonstrations required multiple triggers of the shear to fully sever through tough peduncles. In some instances, the policy would inadvertently reopen the gripper after the peduncle had been successfully severed, resulting in the pepper being dropped.

The final recurring issue occurred with peppers growing close to the ground. When attempting to harvest these low-lying peppers, the robot occasionally pressed them against the ground, leading to collisions and crop damage.

\subsection{Future Works}

Several enhancements could further improve the system’s performance, such as an addition of a dedicated peduncle camera. By incorporating cameras on both sides of the end-effector, the system could gain explicit visual feedback of the peduncle, aiding in accurate targeting of the stem during harvesting. Training a model to explicitly recognize the peduncle and conditioning the diffusion policy on peduncle parameters could further enhance cutting precision.

Addressing occlusions presents another opportunity for improvement. Instead of increasing system complexity with an additional robotic arm, simpler solutions such as an actively controlled leaf blower could effectively clear obstructive foliage. This approach would improve visibility and reduce the risk of missing or damaging peppers, maintaining a balance between system complexity and performance.



Finally, while the existing system has demonstrated promise in pepper harvesting, the methods employed for data collection and policy generation are applicable to a broader range of specialty crop manipulation tasks, such as table grape harvesting, strawberry picking, and grapevine pruning. Furthermore, extending these techniques to bi-manual manipulation tasks could significantly enhance operational dexterity, building on the advancements achieved by UMI \cite{chi2024universal} in dual-arm robotic systems. Such an extension would diversify the applications of automation across various agricultural settings.

\section{Conclusion}

In this study, we presented a system for autonomous robotic pepper harvesting designed to operate in an unprotected outdoor field environment. Leveraging a custom handheld shear-gripper device, we collected 300 pepper harvesting demonstrations, which were used to train a visuomotor policy for robotic harvesting. Our system achieved a harvesting success rate of 28.95\% with a cycle time of 31.71 seconds, demonstrating the feasibility of automated harvesting in complex, unstructured agricultural settings.
As future work, we plan to enhance the system’s robustness by incorporating additional sensors and actuators to address challenges such as peduncle localization and active occlusion removal. Future efforts will also focus on fully automating the process, including ground vehicle navigation and initial positioning of the robotic arm. We hope that this work contributes towards scalable, automated solutions for agricultural harvesting in natural field environments.



\section*{Acknowledgement}

We would like to thank the staff at the ISU Horticultural Research Station, including Ajay Nair, Chad Arnold, Jacie Legois, and Nick Howell, for their support during data collection. We also appreciate the contributions of Warren Whittaker from Carnegie Mellon University and Kalinda Wagner from the University of Pittsburgh. This work used Bridges-2 at Pittsburgh Supercomputing Center \cite{10.1145/3437359.3465593} through allocation \textit{agr240003p} from the Advanced Cyberinfrastructure Coordination Ecosystem: Services \& Support (ACCESS) program, which is supported by National Science Foundation grants \#2138259, \#2138286, \#2138307, \#2137603, and \#2138296. Additionally, this work was supported in part by NSF Robust Intelligence 1956163, and NSF/USDA-NIFA AIIRA AI Research Institute 2021-67021-35329.

\bibliographystyle{ieeetr}
\bibliography{references}

\begin{thebibliography}{10}

\bibitem{machines11010048}
C.~Cheng, J.~Fu, H.~Su, and L.~Ren, ``Recent advancements in agriculture robots: Benefits and challenges,'' {\em Machines}, vol.~11, no.~1, 2023.

\bibitem{mahmoudi11leveraging}
S.~Mahmoudi, A.~Davar, P.~Sohrabipour, R.~B. Bist, Y.~Tao, and D.~Wang, ``Leveraging imitation learning in agricultural robotics: A comprehensive survey and comparative analysis,'' {\em Frontiers in Robotics and AI}, vol.~11, p.~1441312, 2024.

\bibitem{chi2024universal}
C.~Chi, Z.~Xu, C.~Pan, E.~Cousineau, B.~Burchfiel, S.~Feng, R.~Tedrake, and S.~Song, ``Universal manipulation interface: In-the-wild robot teaching without in-the-wild robots,'' in {\em Proceedings of Robotics: Science and Systems (RSS)}, 2024.

\bibitem{chi2023diffusionpolicy}
C.~Chi, S.~Feng, Y.~Du, Z.~Xu, E.~Cousineau, B.~Burchfiel, and S.~Song, ``Diffusion policy: Visuomotor policy learning via action diffusion,'' in {\em Proceedings of Robotics: Science and Systems (RSS)}, 2023.

\bibitem{fountas2020agricultural}
S.~Fountas, N.~Mylonas, I.~Malounas, E.~Rodias, C.~Hellmann~Santos, and E.~Pekkeriet, ``Agricultural robotics for field operations,'' {\em Sensors}, vol.~20, no.~9, p.~2672, 2020.

\bibitem{freeman20233d}
H.~Freeman, E.~Schneider, C.~H. Kim, M.~Lee, and G.~Kantor, ``3d reconstruction-based seed counting of sorghum panicles for agricultural inspection,'' in {\em 2023 IEEE International Conference on Robotics and Automation (ICRA)}, pp.~9594--9600, IEEE, 2023.

\bibitem{10494888}
M.~Lee, A.~Berger, D.~Guri, K.~Zhang, L.~Coffey, G.~Kantor, and O.~Kroemer, ``Towards autonomous crop monitoring: Inserting sensors in cluttered environments,'' {\em IEEE Robotics and Automation Letters}, vol.~9, no.~6, pp.~5150--5157, 2024.

\bibitem{10160650}
C.~H. Kim and G.~Kantor, ``Occlusion reasoning for skeleton extraction of self-occluded tree canopies,'' in {\em 2023 IEEE International Conference on Robotics and Automation (ICRA)}, pp.~9580--9586, 2023.

\bibitem{10611327}
C.~H. Kim, M.~Lee, O.~Kroemer, and G.~Kantor, ``Towards robotic tree manipulation: Leveraging graph representations,'' in {\em 2024 IEEE International Conference on Robotics and Automation (ICRA)}, pp.~11884--11890, 2024.

\bibitem{silwal2022bumblebee}
A.~Silwal, F.~Yandun, A.~K. Nellithimaru, T.~Bates, and G.~Kantor, ``Bumblebee: A path towards fully autonomous robotic vine pruning.,'' {\em Field Robotics}, vol.~2, no.~1, pp.~1661--1696, 2022.

\bibitem{you2022precision}
A.~You, H.~Kolano, N.~Parayil, C.~Grimm, and J.~R. Davidson, ``Precision fruit tree pruning using a learned hybrid vision/interaction controller,'' in {\em 2022 International Conference on Robotics and Automation (ICRA)}, pp.~2280--2286, IEEE, 2022.

\bibitem{xiong2020autonomous}
Y.~Xiong, Y.~Ge, L.~Grimstad, and P.~J. From, ``An autonomous strawberry-harvesting robot: Design, development, integration, and field evaluation,'' {\em Journal of Field Robotics}, vol.~37, no.~2, pp.~202--224, 2020.

\bibitem{7759122}
H.~Yaguchi, K.~Nagahama, T.~Hasegawa, and M.~Inaba, ``Development of an autonomous tomato harvesting robot with rotational plucking gripper,'' in {\em 2016 IEEE/RSJ International Conference on Intelligent Robots and Systems (IROS)}, pp.~652--657, 2016.

\bibitem{sepulveda2020robotic}
D.~Sep{\'u}Lveda, R.~Fern{\'a}ndez, E.~Navas, M.~Armada, and P.~Gonz{\'a}lez-De-Santos, ``Robotic aubergine harvesting using dual-arm manipulation,'' {\em IEEE Access}, vol.~8, pp.~121889--121904, 2020.

\bibitem{lenz2024hortibotadaptivemultiarmrobotic}
C.~Lenz, R.~Menon, M.~Schreiber, M.~P. Jacob, S.~Behnke, and M.~Bennewitz, ``Hortibot: An adaptive multi-arm system for robotic horticulture of sweet peppers,'' 2024.

\bibitem{pan2024development}
Q.~Pan, D.~Wang, J.~Lian, Y.~Dong, and C.~Qiu, ``Development of an automatic sweet pepper harvesting robot and experimental evaluation,'' in {\em 2024 IEEE International Conference on Robotics and Automation (ICRA)}, pp.~15811--15817, IEEE, 2024.

\bibitem{lehnert2016sweet}
C.~Lehnert, I.~Sa, C.~McCool, B.~Upcroft, and T.~Perez, ``Sweet pepper pose detection and grasping for automated crop harvesting,'' in {\em 2016 IEEE international conference on robotics and automation (ICRA)}, pp.~2428--2434, IEEE, 2016.

\bibitem{bac2017performance}
C.~W. Bac, J.~Hemming, B.~Van~Tuijl, R.~Barth, E.~Wais, and E.~J. van Henten, ``Performance evaluation of a harvesting robot for sweet pepper,'' {\em Journal of Field Robotics}, vol.~34, no.~6, pp.~1123--1139, 2017.

\bibitem{lehnert2020performance}
C.~Lehnert, C.~McCool, I.~Sa, and T.~Perez, ``Performance improvements of a sweet pepper harvesting robot in protected cropping environments,'' {\em Journal of Field Robotics}, vol.~37, no.~7, pp.~1197--1223, 2020.

\bibitem{arad2020development}
B.~Arad, J.~Balendonck, R.~Barth, O.~Ben-Shahar, Y.~Edan, T.~Hellstr{\"o}m, J.~Hemming, P.~Kurtser, O.~Ringdahl, T.~Tielen, {\em et~al.}, ``Development of a sweet pepper harvesting robot. j field robot 37: 1027--1039,'' 2020.

\bibitem{porichis2024imitation}
A.~Porichis, M.~Inglezou, N.~Kegkeroglou, V.~Mohan, and P.~Chatzakos, ``Imitation learning from a single demonstration leveraging vector quantization for robotic harvesting,'' {\em Robotics}, vol.~13, no.~7, p.~98, 2024.

\bibitem{9013082}
K.~Motokura, M.~Takahashi, M.~Ewerton, and J.~Peters, ``Plucking motions for tea harvesting robots using probabilistic movement primitives,'' {\em IEEE Robotics and Automation Letters}, vol.~5, no.~2, pp.~3275--3282, 2020.

\bibitem{paraschos2013probabilistic}
A.~Paraschos, C.~Daniel, J.~R. Peters, and G.~Neumann, ``Probabilistic movement primitives,'' {\em Advances in neural information processing systems}, vol.~26, 2013.

\bibitem{van2024using}
R.~van~de Ven, A.~L. Shoushtari, A.~Nieuwenhuizen, G.~Kootstra, and E.~J. van Henten, ``Using learning from demonstration (lfd) to perform the complete apple harvesting task,'' {\em Computers and Electronics in Agriculture}, vol.~224, p.~109195, 2024.

\bibitem{van2024duallqr}
R.~van~de Ven, A.~Nieuwenhuizen, E.~J. van Henten, and G.~Kootstra, ``Duallqr: Efficient grasping of oscillating apples using task parameterized learning from demonstration,'' {\em arXiv preprint arXiv:2409.16957}, 2024.

\bibitem{misimi2018robotic}
E.~Misimi, A.~Olofsson, A.~Eilertsen, E.~R. {\O}ye, and J.~R. Mathiassen, ``Robotic handling of compliant food objects by robust learning from demonstration,'' in {\em 2018 IEEE/RSJ International Conference on Intelligent Robots and Systems (IROS)}, pp.~6972--6979, IEEE, 2018.

\bibitem{ho2020denoising}
J.~Ho, A.~Jain, and P.~Abbeel, ``Denoising diffusion probabilistic models,'' {\em Advances in neural information processing systems}, vol.~33, pp.~6840--6851, 2020.

\bibitem{fang2019survey}
B.~Fang, S.~Jia, D.~Guo, M.~Xu, S.~Wen, and F.~Sun, ``Survey of imitation learning for robotic manipulation,'' {\em International Journal of Intelligent Robotics and Applications}, vol.~3, pp.~362--369, 2019.

\bibitem{campos2021orb}
C.~Campos, R.~Elvira, J.~J.~G. Rodr{\'\i}guez, J.~M. Montiel, and J.~D. Tard{\'o}s, ``Orb-slam3: An accurate open-source library for visual, visual--inertial, and multimap slam,'' {\em IEEE Transactions on Robotics}, vol.~37, no.~6, pp.~1874--1890, 2021.

\bibitem{10.1145/3437359.3465593}
S.~T. Brown, P.~Buitrago, E.~Hanna, S.~Sanielevici, R.~Scibek, and N.~A. Nystrom, ``Bridges-2: A platform for rapidly-evolving and data intensive research,'' in {\em Practice and Experience in Advanced Research Computing 2021: Evolution Across All Dimensions}, PEARC '21, (New York, NY, USA), Association for Computing Machinery, 2021.

\end{thebibliography}

\end{document}